\documentclass[conference]{IEEEtran}

\usepackage{cite}
\usepackage{amsmath,amssymb,amsfonts}
\usepackage{algorithmic}
\usepackage{graphicx}
\usepackage{textcomp}
\usepackage{xcolor}

\usepackage{multirow,tabularx}
\usepackage{url}
\newcolumntype{C}{>{\centering\arraybackslash}m{1.5cm}}
\newcolumntype{M}{>{\centering\arraybackslash}X}
\newcommand{\nb}[1]{\noindent \textbf{#1}}

\makeatletter
\newcommand{\linebreakand}{%
  \end{@IEEEauthorhalign}
  \hfill\mbox{}\par
  \mbox{}\hfill\begin{@IEEEauthorhalign}
}
\makeatother

\def\BibTeX{{\rm B\kern-.05em{\sc i\kern-.025em b}\kern-.08em
    T\kern-.1667em\lower.7ex\hbox{E}\kern-.125emX}}
\begin{document}

\title{Holistic Visual-Textual Sentiment Analysis with Prior Models}

\author{
\IEEEauthorblockN{Junyu Chen}
\IEEEauthorblockA{\textit{Department of Computer Science} \\
\textit{University of Rochester}\\
Rochester, USA \\
jchen175@ur.rochester.edu}
\and
\IEEEauthorblockN{Jie An}
\IEEEauthorblockA{\textit{Department of Computer Science} \\
\textit{University of Rochester}\\
Rochester, USA \\
jan6@cs.rochester.edu}
\and
\IEEEauthorblockN{Hanjia Lyu}
\IEEEauthorblockA{\textit{Department of Computer Science} \\
\textit{University of Rochester}\\
Rochester, USA \\
hlyu5@ur.rochester.edu}
\linebreakand
\IEEEauthorblockN{Christopher Kanan}
\IEEEauthorblockA{\textit{Department of Computer Science} \\
\textit{University of Rochester}\\
Rochester, USA \\
ckanan@cs.rochester.edu}
\and
\IEEEauthorblockN{Jiebo	Luo}
\IEEEauthorblockA{\textit{Department of Computer Science} \\
\textit{University of Rochester}\\
Rochester, USA \\
jluo@cs.rochester.edu}
}

\maketitle

\begin{abstract}
Visual-textual sentiment analysis aims to predict sentiment with the input of a pair of image and text, which poses a challenge in learning effective features for diverse input images. To address this, we propose a holistic method that achieves robust visual-textual sentiment analysis by exploiting a rich set of powerful pre-trained visual and textual prior models. The proposed method consists of four parts: (1) a visual-textual branch to learn features directly from data for sentiment analysis, (2) a visual expert branch with a set of pre-trained ``expert'' encoders to extract selected semantic visual features, (3) a CLIP branch to implicitly model visual-textual correspondence, and (4) a multimodal feature fusion network based on BERT to fuse multimodal features and make sentiment predictions. Extensive experiments on three datasets show that our method produces better visual-textual sentiment analysis performance than existing methods.
\end{abstract}

\begin{IEEEkeywords}
visual-textual sentiment analysis, multimodal fusion, deep neural network.
\end{IEEEkeywords}

\section{Introduction}
\begin{figure}[t]
\centering
\includegraphics[width=0.47\textwidth]{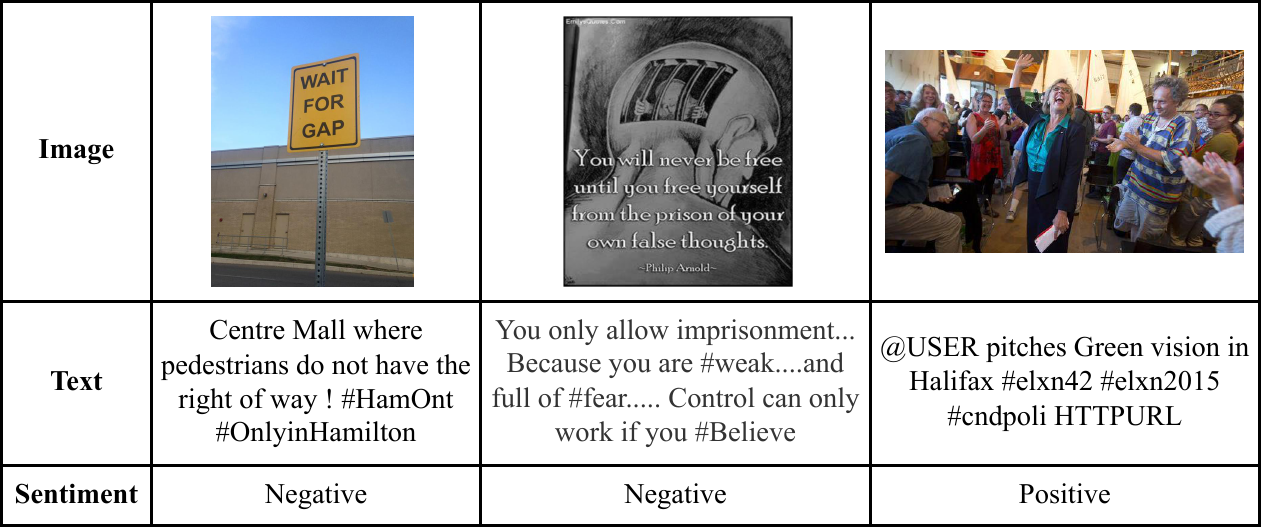}
\caption{Examples from a visual-textual sentiment dataset, where image and text are used to jointly express the sentiment. Clearly, it is challenging to predict the sentiment only from the image in some cases.}
\label{fig:intro}
\end{figure}

Visual-textual sentiment analysis~\cite{you2016robust} lies at the intersection of computer vision and natural language processing. Given a pair of image and text, the visual-textual sentiment analysis task aims at recognizing polarized sentiments (i.e., positive, negative, and neutral) or fine-grained emotions (e.g., happy, angry, calm, etc.). A typical application is to predict sentiments in social media posts where users post text and images to jointly express their feelings (Fig.~\ref{fig:intro}).

The key to visual-textual sentiment analysis is extracting useful visual/textual features for sentiment inference. Existing studies~\cite{you2016robust,xu2023multimodal,you2016cross} typically employ two encoders for visual and textual information extraction, with the visual encoder based on CNN and the text encoder based on either RNN or Transformer. A feature fusion module then combines visual and textual features for sentiment inference.
Although this framework has made significant progress, a main challenge remains: it is difficult to effectively extract useful visual features and recognize the {\it abstract} sentiment of images. Sentiment analysis requires a relatively higher level of information, which may be represented by scene, object, facial expression, or their combinations~\cite{yuan2013sentribute,zheng2017saliency}, which is \textbf{difficult to learn with only one network without any \textit{prior knowledge}}. 
Furthermore, images in visual-textual sentiment datasets can come from  \textbf{highly diverse domains}. For example, as shown in Fig.~\ref{fig:intro}, some images are screenshots of text. Without an understanding of the embedded image text, it is unlikely to infer the sentiment accurately. Therefore, a powerful visual feature extractor that can capture various effective information from images and handle the heterogeneity of various domains of images is desired for visual-textual sentiment analysis.

\begin{figure*}[t]
\centering
\includegraphics[width=0.99\textwidth]{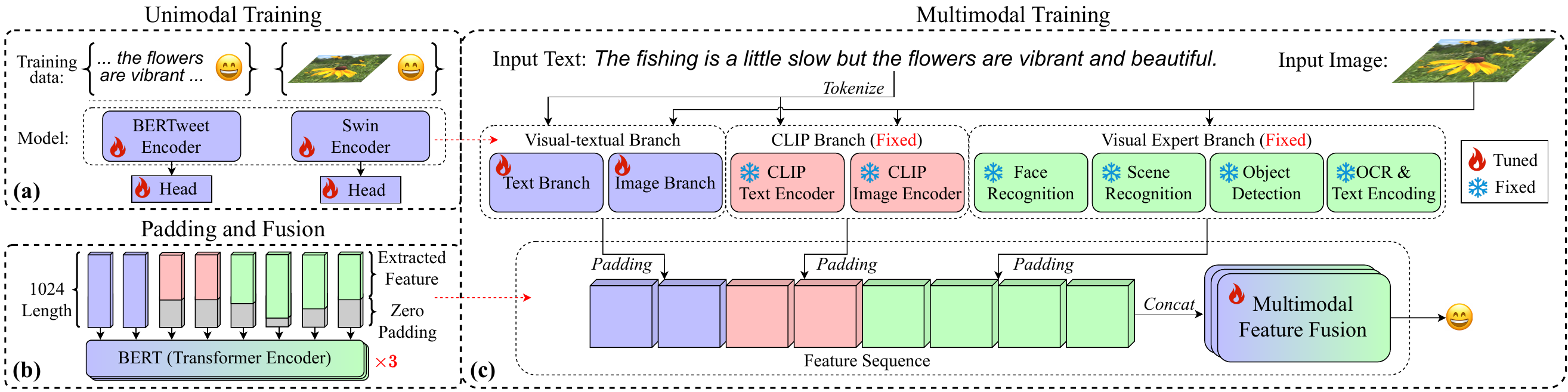}
\vspace{-4mm}
\caption{Overview of the proposed method’s architecture and training procedure. Our framework (\textbf{c}) consists of four parts: (1) A visual-textual branch to learn visual and textual features for sentiment prediction, (2) A visual expert branch to equip the method with a strong visual prior, (3) A CLIP branch to implicitly model the visual-textual correspondence with aligned embeddings, and (4) A multimodal feature fusion module to integrate all information and make holistic sentiment predictions. Initial training (\textbf{a}) fine-tunes the visual-textual branch separately on unimodal data to capture sentiment features before proceeding to multimodal training (\textbf{b}, \textbf{c}). The dataset splits remain consistent throughout the two training phases.
}
\vspace{-6mm}
\label{fig:framework}
\end{figure*}

In this study, we propose a holistic approach for visual-textual sentiment analysis, aiming for robust performance by leveraging the power of pre-trained feature extractors. The proposed algorithm, named \textbf{VSA-PF} (\textbf{V}isual-textual \textbf{S}entiment \textbf{A}nalysis with \textbf{P}re-trained \textbf{F}eatures; Fig.~\ref{fig:framework}), consists of four parts. First, a visual-textual branch based on the Swin Transformer~\cite{liu2021swin} and BERTweet~\cite{bertweet} is used to learn visual and textual features for sentiment prediction. It captures the important information for sentiment prediction directly from the dataset. These learnable encoders are pre-trained on large-scale datasets for generalizability.

Second, a visual expert branch with a rich set of pre-trained ``expert'' visual encoders is used. These encoders extract semantic visual features, such as face~\cite{zhang2016joint}, object~\cite{Jocher_YOLOv5_by_Ultralytics_2020}, scene~\cite{zhou2017places}, and OCR, which are useful for sentiment analysis of {\bf different visual contents}~\cite{yuan2013sentribute,xu2017multisentinet,cheema2021fair} but hard to learn directly from the visual-textual sentiment analysis dataset. 

Third, we adopt a CLIP branch. Since CLIP is trained with a large number of image-text pairs, on one hand, it improves the generalizability of the proposed method to handle {\it unseen or weakly} sentiment images. On the other hand, because CLIP is trained to minimize the cosine distance between paired images and textual captions, image-text correspondence can be implicitly induced with the extracted visual and textual features from the CLIP encoder. 

Finally, a multimodal feature fusion network based on BERT~\cite{devlin2018bert} (i.e., Transformer encoder) is used to fuse multi-domain multimodal features and make sentiment predictions, which facilitates effective modeling of the complex interactions between different features. Extensive experiments on three datasets demonstrate that the proposed method outperforms state-of-the-art frameworks on visual-textual sentiment analysis. Furthermore, we conduct an ablation study to demonstrate the necessity of the visual-textual branch, show the contribution of each pre-trained expert feature, and compare the different architecture choices of the fusion network.

\textbf{Our contributions are three folds}: 
($1$) Utilizing the pre-trained CLIP model, we extract \textit{aligned visual and textual features}, enhancing sentiment prediction by implicitly modeling visual-textual correspondence. 
($2$) Incorporating a set of ``expert'' visual encoders pre-trained for various visual tasks, our method benefits from a \textbf{strong visual prior}, capturing \textit{complicated, diverse, and nuanced} visual information for sentiment analysis.
($3$) Our holistic framework, composed of CLIP, visual-expert, and trained visual-textual branches, achieves better performance than the state-of-the-art methods on three existing datasets.

\section{Related Work}
\nb{Textual Sentiment Analysis.} 
Textual sentiment analysis is a well-established task in NLP. Traditional methods such as VADER~\cite{hutto2014vader} typically rely on hand-crafted and lexicon-based features. In recent years, deep learning techniques have been widely adopted. For instance, \cite{wang2016dimensional} proposed using a combination of CNN and LSTM to capture both local affective information and long-distance sentence dependencies. More recently, transformer-based models~\cite{devlin2018bert} such as BERTweet~\cite{bertweet} have set new benchmarks in a variety of NLP tasks, including sentiment analysis.

\nb{Visual Sentiment Analysis.}
Visual sentiment analysis aims to explore the sentiment evoked or expressed by images. Traditional methods often rely on low- to mid-level features such as color distribution~\cite{yuan2013sentribute} and adjective-noun-pair detector~\cite{borth2013large}. Many recent approaches are based on CNNs to take advantage of the expressive power of pre-training. For example, \cite{you2015robust} designed a progressive fine-tuning procedure for enhanced transfer learning performance. \cite{you2017visual} adopted attention modules with CNNs to highlight sentiment-revealing image regions. Compared with textual data, visual sentiment analysis poses \textbf{a unique challenge due to the high level of visual diversity, semantic abstraction, and viewer subjectivity}. Some approaches attempt to address these by incorporating extra features~\cite{saito2023visual}, such as utilizing facial expression recognition to improve sentiment prediction in images with faces~\cite{yuan2013sentribute,ge2013is}.

\nb{Visual-textual Sentiment Analysis.}
Visual-textual sentiment analysis integrates information from both visual and textual modalities to predict sentiment.
Some approaches in this field focus on fusing multimodal information effectively~\cite{you2016cross,xu2023multimodal}. Other works concentrate on extracting effective features from each modality~\cite{li2022clmlf,xiao2022adaptive,dai2018integrating}.
For example, \cite{li2022clmlf} used pretext tasks such as masked language/region modeling to help the model capture sentiment-related features. \cite{xu2017multisentinet} and~\cite{yang2020image} adopted object and scene features of images. 
The most relevant work to ours is Se-MLNN~\cite{cheema2021fair}, which adopted pre-trained encoders to extract visual features including objects, scenes, facial expressions, and CLIP embeddings. However,   \textbf{our work differs in several major ways} from \cite{cheema2021fair}. First, we incorporate a \textbf{learnable} branch to directly learn important multimodal features from the training dataset, in contrast to Se-MLNN's reliance on pre-trained and fixed feature extractors. 
Second, we employ a pre-trained OCR encoder to explicitly handle texts in images, which is very prevalent in all the studied datasets (but largely overlooked by existing methods). This OCR feature proves important for accuracy in our experiments. 
Additionally, we use BERT to fuse different features, which is more effective at learning the interactions between multimodal multi-domain features. Overall, our experiments demonstrate that our holistic method outperforms Se-MLNN in all three public datasets.

\section{Method}
Our framework, illustrated in Fig.~\ref{fig:framework}, integrates four primary components: (1) a trainable visual-textual branch to directly extract sentiment features from the dataset, (2) a visual expert branch consisting of four visual feature extractors to capture semantic visual information from open-domain images, (3) a CLIP branch for both image and text inputs to implicitly model the cross-modal correspondence, and (4) a multimodal fusion network to fuse all information and make holistic sentiment prediction. This section details each component.
\subsection{Visual-textual Branch} 
The visual-textual branch aims at extracting sentiment-related features directly from the datasets. Two networks are adopted to extract features from text and image, respectively. 

\nb{Text Encoder.} We employ BERTweet-large~\cite{bertweet} as the text encoder. BERTweet is based on BERT architecture~\cite{devlin2018bert} and was pre-trained using $873$M English tweet. It achieves state-of-the-art performance in tweets classification tasks. Since the text of visual-textual sentiment datasets originated from social media which is similar to tweets, the text encoder can acquire a strong prior from pre-training.
We fine-tune the model with the text of the training set to adapt it to sentiment analysis. Next, we remove the classification head of the BERTweet model, and utilize the last layer's hidden state of the classification token as the text representation, which is an embedding of $1024$ dimensions.

\nb{Image Encoder.} We use Swin Transformer~\cite{liu2021swin} as the image encoder, specifically a Swin-base network pre-trained on ImageNet-21k at a resolution of $384\times384$. This pre-training enables the model to more robustly handle diverse and noisy open-domain images in the datasets. Next, similar to the text encoder, we fine-tune the Swin-base model with only images of the datasets, remove the classification head, and use the last hidden-state layer's average pooling to generate a $1024$-dimensional image embedding.
\subsection{Visual Expert Branch} 
The visual expert branch employs four pre-trained neural networks to selectively extract high-valued face, scene, object, and optical character recognition (OCR) features in images.

\nb{Face Recognition.} 
Face images are of pivotal importance in sentiment analysis~\cite{yuan2013sentribute,cheema2021fair}. As shown in Table~\ref{table:datasets summary}, portrait photos with faces account for over $40\%$ of images in the datasets. To extract effective face features, we first use the MTCNN face detector~\cite{zhang2016joint} to filter out images without faces, then identify and retain the largest face in each image. We then adopt a pre-trained Facenet~\cite{schroff2015facenet} based on Inception-ResNet architecture as the face feature extractor, which was pre-trained on VGGFace2. 
The resulting $512$-dimensional embedding represents the face features.

\nb{Scene and Object Detection.} Scenes and objects are pivotal in conveying sentiment in images~\cite{borth2013large,yang2021multimodal}. For example, a photo of a delicious meal in a fancy restaurant reveals enjoyment, while a solitary tree in a graveyard implies melancholy. To capture these elements, our visual expert branch employs YOLOv5~\cite{Jocher_YOLOv5_by_Ultralytics_2020} pre-trained on the MS COCO dataset for object detection. The output, i.e., the predicted logit for each category, is summed up into an $80$-dimensional vector to represent the number and types of objects in each image.
For scene detection, we use a DenseNet161 pre-trained on the Place365 dataset~\cite{zhou2017places}. This model can recognize $365$ types of scenes, such as ``airfield '', ``campus'' and ``highway''. We use the output logit of each image as the scene features, which are $365$-dimensional.
    
\nb{OCR.} Text within images, such as internet memes, event posters, and screenshots of messages, is prevalent on social media platforms like Twitter. To extract OCR features, we employ a pre-trained Sentence-Transformer based on MPNet~\cite{reimers-2019-sentence-bert} as our OCR encoder. The process begins with extracting English words from images using Tesseract-OCR Engine~\cite{pyocr}. These words are then concatenated into a sequence and processed through the Sentence-Transformer. The output $768$-dimensional sentence embeddings serve as the OCR features. To ensure relevance, we only extract OCR features from images containing at least five words, minimizing the impact of trivial text.
\subsection{CLIP Branch} 
The CLIP branch leverages CLIP's pre-trained image and text encoders to extract corresponding features. The image encoder is a $24$-layer, $16$-head vision transformer, while the text encoder is a $12$-layer, $12$-head transformer, both with an embedding dimension of $768$. CLIP~\cite{radford2021learning} was pre-trained on a massive dataset of $400$ million image-text pairs using a contrastive loss to minimize the distance between the paired text and image. This training enables CLIP's encoders to discern high-level correlations between images and text that unimodal feature extractors cannot capture.

\subsection{Multimodal Feature Fusion} 
We employ a $3$-layer, $512$-width BERT model (i.e., Transformer encoder) with $4$ attention heads to integrate all visual and textual features for sentiment prediction. As illustrated in Fig.~\ref{fig:framework}b, we first zero pad all the extracted features to the same size of $1024$ to prevent distortion before fusion. These features are arranged into a sequence and processed by the BERT model. This process effectively fuses the information from the visual-textual, visual expert, and CLIP branches to make the final sentiment prediction.

\section{Experiments}
To demonstrate the performance of the proposed method, we conduct experiments on three public datasets. In this section, we introduce the experimental setup and show the experiment results along with key findings. The source code and the dataset splits used in our experiments are available at \url{https://github.com/jchen175/VSA-PF}.
\subsection{Datasets and Preprocessing}
\nb{MVSA-Single and MVSA-Multiple.} 
These datasets contain $4,869$ and $19,600$ image-text pairs from Twitter. Each pair is assigned a human-annotated sentiment label (positive, negative, neutral). 
We follow the preprocessing procedure of~\cite{xu2017multisentinet} to remove pairs with conflicting sentiments in two modalities or disagreement among annotators. 
This results in $4,511$ and $17,025$ pairs for MVSA-Single and MVSA-Multiple, respectively.

\nb{TumEmo.} TumEmo~\cite{yang2020image} is a large visual-textual emotion dataset containing $195,265$ image-text pairs from Tumblr, classified into seven emotion classes (i.e., angry, bored, calm, fearful, happy, loving, and sad) if the hashtags in the text contain certain keywords related to that emotion class. We adhere to the original study's methodology, removing all hashtags that could reveal the ground truth label.

We follow BERTweet's normalization procedure for text preprocessing, where emojis, user mentions, and URLs are replaced with special tokens, and the use of punctuation and abbreviations is standardized across sentences. Table~\ref{table:datasets summary} shows the detailed statistics of the datasets after preprocessing.
For each dataset, we adopt an 8:1:1 split ratio for the training, validation, and test sets in a stratified manner, where the distributions of annotation labels are consistent across these sets. 
Pre-trained features are extracted once and saved for reuse. 

\begin{table}[t]
\tiny
\caption{The summary statistics of the datasets. We show the number of samples belonging to each sentiment/emotion category and the ratio (\%) of samples that contain faces, objects, or \textbf{image text} in three datasets. Note that \textbf{the percentage of image text is quite high} across all three datasets.} 
\centering
\resizebox{\linewidth}{!}{
    \begin{tabular}{ccc|ccc}
    \hline
    \textbf{Dataset} & \textbf{Sentiment} & \textbf{Samples} & \textbf{Face} & \textbf{Object} & \textbf{OCR} \\
    \hline
    \multirow{4}{*}{\shortstack{MVSA-\\Single}}
        & Positive & 2,683 & 55.65 & 86.92 & 14.39 \\
        & Neutral & 470   & 22.98 & 73.40 & 18.30 \\
        & Negative & 1,358 & 46.32 & 82.11 & 17.89 \\
          & \textbf{All} & 4,511 & 49.43 & 84.06 & 15.85 \\
    \hline
    \multirow{4}{*}{\shortstack{MVSA-\\Multiple}}
        & Positive & 11,318 & 50.14 & 84.94 & 20.81 \\
        & Neutral & 4,408 & 45.30 & 79.81 & 31.22 \\
        & Negative & 1,299 & 47.04 & 79.37 & 28.41 \\
        & \textbf{All} & 17,025 & 48.65 & 83.19 & 24.08 \\
    \hline
    \multirow{8}{*}{TumEmo} 
          & Angry & 14,554 & 44.32 & 73.01 & 21.82 \\
          & Bored & 32,283 & 51.86 & 83.79 & 12.32 \\
          & Calm & 18,109 & 18.56 & 60.82 & 18.82 \\
          & Fearful & 20,264 & 24.37 & 59.64 & 42.42 \\
          & Happy & 50,267 & 49.12 & 80.57 & 19.81 \\
          & Loving & 34,511 & 42.31 & 74.71 & 26.42 \\
          & Sad & 25,277 & 35.18 & 67.10 & 40.54 \\
          & \textbf{All} & 195,265 & 40.81 & 73.76 & 24.83 \\
    \hline
    \end{tabular}%
    }
\vspace{-5mm}
\label{table:datasets summary}
\end{table}

\begin{table}[h]
\centering
\small
\caption{Quantitative results on three datasets. We show reported baseline results in their papers and compare our method with theirs using our dataset splits (rep). The results of two MVSA datasets are averaged over 10 folds. The best performance is in bold and the second best is underlined.}

\resizebox{\linewidth}{!}{
    \begin{tabular}{c|cc|cc|cc}
    \hline
     \multicolumn{1}{c|}{\multirow{2}{*}{\textbf{Model}}} & \multicolumn{2}{c|}{\textbf{MVSA-Single}} & \multicolumn{2}{c|}{\textbf{MVSA-Multiple}} & \multicolumn{2}{c}{\textbf{TumEmo}} \\
          \multicolumn{1}{c|}{} & \textbf{Acc$\uparrow$} & \textbf{F1$\uparrow$} & \textbf{Acc$\uparrow$} & \textbf{F1$\uparrow$} & \textbf{Acc$\uparrow$} & \textbf{F1$\uparrow$} \\
    \hline
        MGNNS~\cite{yang2021multimodal} & 73.77 & 72.70 & 72.49 & 69.34 & 66.72 & 66.69 \\
        Se-MLNN~\cite{cheema2021fair} & 75.33 & 73.76 & 66.35 & 61.89 & - & - \\
        CLMLF~\cite{li2022clmlf} & 75.33 & 73.46 & 72.00 & 69.83 & - & - \\\hline
        MGNNS (rep) & 71.04 & 69.68 & 69.43 & 65.88 & 65.33 & 65.02 \\
        Se-MLNN (rep) & \underline{73.09} & 71.31 & 69.22 & 64.57 & 58.16 & 57.99 \\
        CLMLF (rep) & 72.73 & \underline{71.62} & \underline{70.08} & \underline{66.33} & \underline{71.12} & \underline{71.01} \\
        VSA-PF (ours) & \textbf{75.58} & \textbf{74.48} & \textbf{71.26} & \textbf{68.72} & \textbf{76.58} & \textbf{76.57} \\ 
          \hline
    \end{tabular}
}
\vspace{-4.5mm}

\label{table:exp results}
\end{table}
\subsection{Model Training}
In model training, the CLIP and visual expert branches are fixed. As shown in Fig.~\ref{fig:framework}, our training procedure consists of two steps.

\nb{Unimodal Training.} 
We first train the visual-textual branch separately with either text or images in the datasets. This procedure enables the encoders to learn enough knowledge for sentiment analysis from unimodal data. 
For the visual encoder (Swin Transformer), we train exclusively with images for $15$ epochs, with a learning rate of $5\mathrm{e}{-5}$ for MVSA and $1\mathrm{e}{-4}$ for TumEmo, and batch sizes of $16$ for MVSA and $32$ for TumEmo. For the textual encoder (BERTweet), we train with only text for $20$ epochs, setting the learning rate to $8\mathrm{e}{-6}$ for MVSA and $5\mathrm{e}{-5}$ for TumEmo, and batch sizes of $16$ for MVSA and $64$ for TumEmo.

\nb{Multimodal Training.} 
After unimodal training, we train the entire model with both images and text. Note the dataset splits remain the same in the two stages. We initialize the visual-textual branch with the best-performing parameters (those with the lowest loss on the validation set) in unimodal training. Next, we train the visual-textual branch and the final multimodal feature fusion module together end-to-end. We use the AdamW optimizer with learning rates of $5\mathrm{e}{-6}$ for MVSA and $1\mathrm{e}{-5}$ for TumEmo, a batch size of $16$ for all datasets, and a dropout rate of $0.5$ to mitigate over-fitting. The training lasts $30$ epochs, and the parameters that yield the highest accuracy on the validation set are then used for testing.
\subsection{Baselines}
We benchmark our method against three publicly available, state-of-the-art methods in visual-textual sentiment analysis:

\noindent\textbf{MGNNS}~\cite{yang2021multimodal} introduces a multi-channel graph neural network to model object, scene, and text representations based on the global co-occurrence patterns across the dataset.

\noindent\textbf{Se-MLNN}~\cite{cheema2021fair} integrates several pre-trained image features with contextual text features extracted via RoBERTa to comprehensively represent multimodal contents.

\noindent\textbf{CLMLF}~\cite{li2022clmlf} introduces a contrastive learning framework to enable effective learning of generalized representations for both image and text in visual-textual sentiment analysis.

Among the compared methods, Se-MLNN relies only on pre-trained features and is lightweight with 1M parameters, while our method and compared CLMLF, MGNNS have comparable sizes (500M, 200M, 100M respectively).

\subsection{Quantitative Comparison}
Table~\ref{table:exp results} shows the quantitative comparison between our proposed method and established baselines. 
While baseline results from their respective papers vary due to different dataset splits, we ensure a fair comparison by training each model with our dataset splits and reporting the metrics. Results for the MVSA datasets are averaged over 10 folds. 

On the MVSA and TumEmo datasets, the proposed method outperforms other strong baselines, highlighting its effectiveness.
The proposed visual-textual branch is based on pre-trained Swin Transformer and BERTweet, which allow our framework to employ the power of advanced neural network architectures. In contrast, Se-MLNN relies solely on pre-trained features \textbf{without learning useful sentiment-related features directly from the dataset}, which may lead to its compromised performance on the more complex TumEmo dataset with fine-grained emotion annotations. 
Regarding the visual expert branch, our method has four different visual features while OCR and face features are not used by MGNNS~\cite{yang2021multimodal}. It is hard for a trainable visual encoder
to learn these missing visual features directly.

 \subsection{Ablation Study}
 In our ablation study, we assess the contribution of each pre-trained and learned feature, the significance of multimodal input, and compare different fusion network architectures.

\begin{table}[t]
\normalsize
\caption{Ablation study results. We show the performance changes on accuracy and weighted-F1 when: (1) removing an encoder in our framework, (2) removing the fine-tuning procedure of the visual-textual branch, or (3) replacing the fusion network with MLP. 
}
\vspace{-2mm}
\centering
\resizebox{\linewidth}{!}{
\begin{tabular}{c|cc|cc|cc}
    \hline
     \multicolumn{1}{c|}{\multirow{2}{*}{\textbf{Model}}} & \multicolumn{2}{c|}{\textbf{MVSA-Single}} & \multicolumn{2}{c|}{\textbf{MVSA-Multiple}} & \multicolumn{2}{c}{\textbf{TumEmo}} \\
          \multicolumn{1}{c|}{} & \textbf{Acc$\uparrow$} & \textbf{F1$\uparrow$} & \textbf{Acc$\uparrow$} & \textbf{F1$\uparrow$} & \textbf{Acc$\uparrow$} & \textbf{F1$\uparrow$} \\
    \hline
    VSA-PF (ours) & 75.58 & 74.48 & 71.26 & 68.72 & 76.58 & 76.57 \\\hline
    VSA-PF (using MLP) & -0.52 & -0.77 & -0.22 & -0.76 & +0.09 & +0.09 \\\hline
    w/o Swin & -2.91 & -3.39 & -0.62 & -1.09 & -0.34 & -0.31 \\
    w/o BERTweet & -4.70 & -5.31 & -2.41 & -4.02 & -6.68 & -6.64 \\
    w/o Fine-tuning & -1.23 & -1.21 & -1.08 & -1.53 & -0.20 & -0.19 \\\hline
    w/o CLIP-Image & -0.42 & +0.04 & -0.31 & -0.67 & -0.70 & -0.72 \\
    w/o CLIP-Text & -0.46 & -0.42 & -0.33 & -0.74 & -0.18 & -0.19 \\
    w/o Face & -0.82 & -0.94 & -0.38 & -0.73 & -0.19 & -0.21 \\
    w/o OCR & -0.93 & -1.42 & -0.42 & -0.80 & -0.05 & -0.06 \\
    w/o Scene & -0.34 & -0.41 & -0.27 & -0.61 & -0.02 & -0.04 \\
    w/o Object & -0.68 & -0.69 & -0.40 & -0.66 & -0.14 & -0.17 \\
    \hline
    \end{tabular}
    }
\vspace{-6mm}
    
\label{table:ablation branch}
\end{table}

\begin{figure*}[t]
\centering
\includegraphics[width=0.88\textwidth]{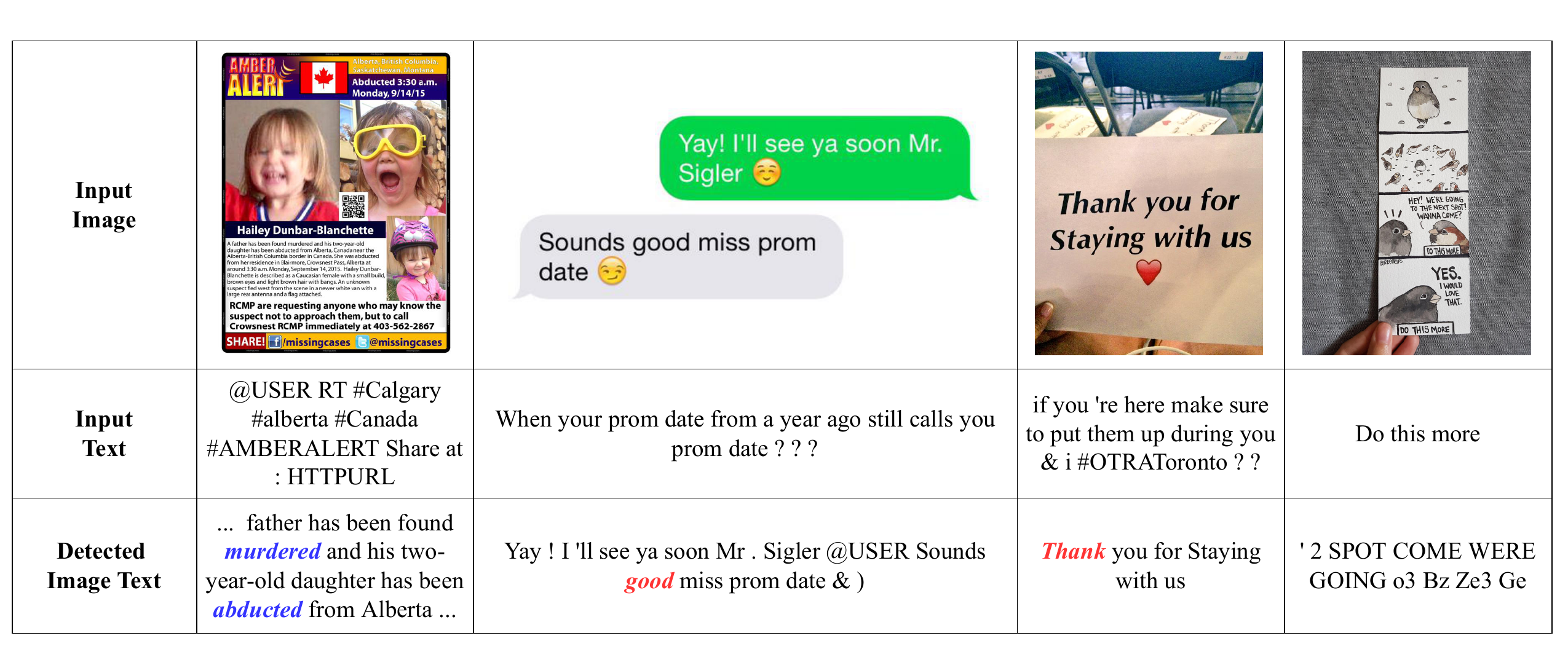}
\vspace{-6mm}
\caption{Examples from visual-textual sentiment datasets, where the sentiment is mainly revealed by image text. We give three instances where the OCR engine accurately detected the image text and helped improve the accuracy, along with one failure case in which the OCR engine failed to detect hand-written image text.
}
\label{fig:OCR}
\vspace{-2mm}
\end{figure*}

\begin{figure*}[t]
\centering
\includegraphics[width=0.88\textwidth]{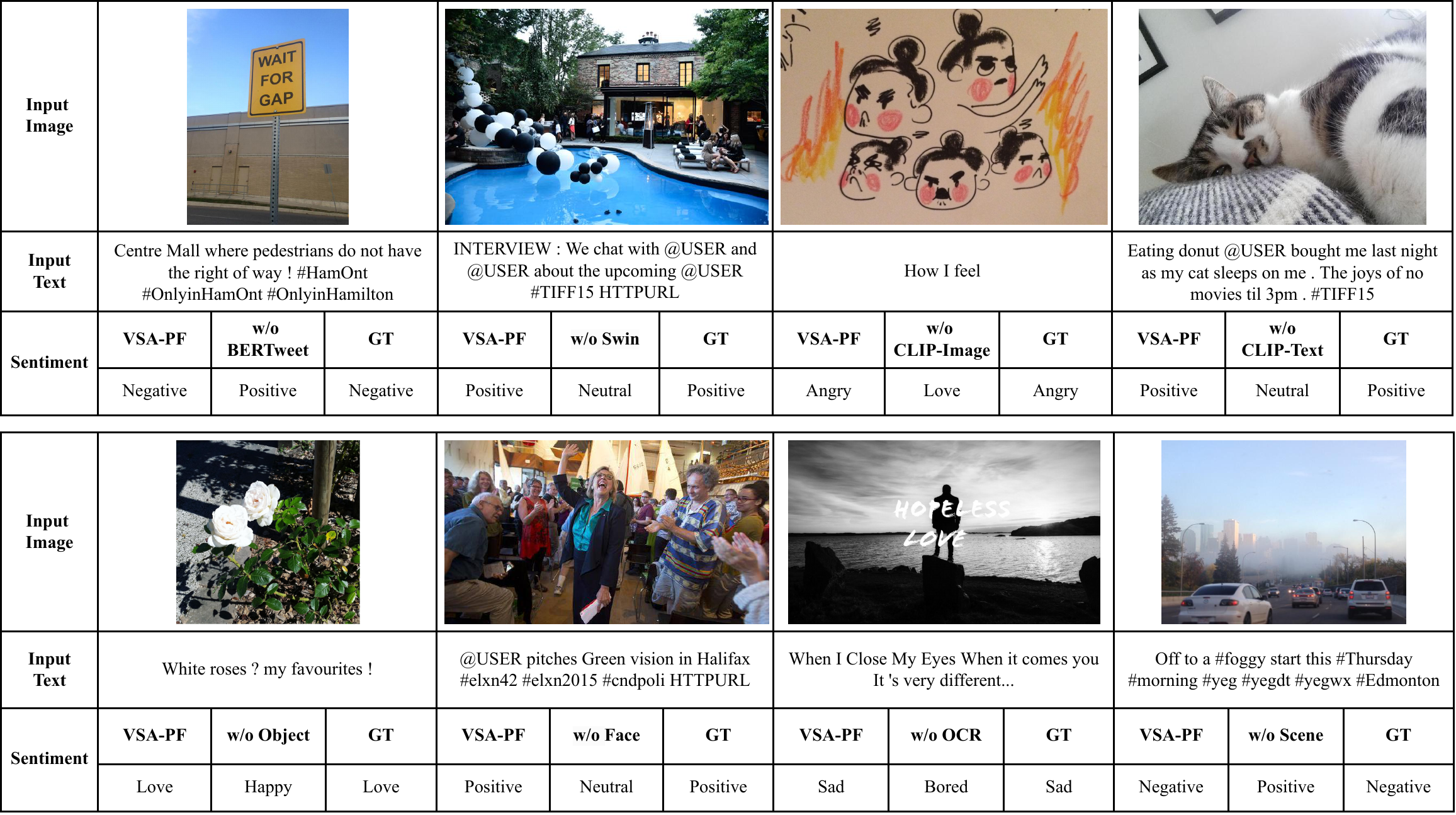}
\vspace{-4mm}
\caption{Examples from visual-textual sentiment datasets. Each column represents a branch and a corresponding data pair, where our VSA-PF model accurately predicts sentiments for these examples, but the removal of the branch results in incorrect prediction for the associated sample.
}
\vspace{-4mm}
\label{fig:qualitative}
\end{figure*}

 \nb{The Importance of Different Branches.}
We analyze performance changes when individual encoders are removed. Additionally, we provide a qualitative illustration (Fig.~\ref{fig:qualitative}), showcasing how each branch contributes to accurate sentiment predictions.
As detailed in Table~\ref{table:ablation branch},
the most substantial decline occurs with the removal of BERTweet, underscoring the critical role of textual information in visual-textual sentiment analysis. We hypothesize the reason is that learning sentiment from text is easier than from images, since adjectives in text can convey sentiment more directly.
In the MVSA datasets, the removal of Swin Transformer, or the fine-tuning of the visual-textual branch (i.e., using fixed pre-trained Swin and BERTweet) leads to a notable performance decrease. 
This indicates the significance of directly learning sentiment-related features from the datasets. Among visual expert features, removing OCR causes the largest performance drop. This is in line with our observation about the prevalence and importance of image text in social media, which has been ignored in previous studies. We show some example cases in Fig.~\ref{fig:OCR}. When the input image contains text to deliver sentiment-related information, the OCR feature can indeed help make correct sentiment predictions. In general, removing any of the pre-trained features diminishes performance, highlighting the necessity of each encoder. We present examples in Fig.~\ref{fig:qualitative} to illustrate the importance of these encoders, where their removal leads to incorrect results. Removing the CLIP image encoder results in a decrease in accuracy and a marginal increase in weighted F1 on MVSA-Single. This could probably be caused by the class imbalance present in the dataset, as shown in Table~\ref{table:datasets summary}.
For TumEmo, the removal of the CLIP image encoder significantly reduces performance. This suggests that the CLIP encoder captures unique image features distinct from the learnable Swin Transformer and is particularly effective in handling the noisy, open-domain images in this complex dataset.

In the ablation study on TumEmo, we find that the OCR and scene features only marginally help sentiment analysis. This observation aligns with \cite{yang2021multimodal}’s ablation study, where scene features had a lesser impact on TumEmo compared to MVSA. A possible explanation is that the OCR and scene encoders might not be sufficiently robust for the complex task of seven-class emotion classification: (1) OCR may not detect all the text present in the images (last column in Fig.~\ref{fig:OCR}), (2) the text information in images may be inconsistent with the input text, and (3) it is more difficult to infer seven-class emotional predictions from scene features. 
\begin{table}[t]
\normalsize
\caption{Ablation study results. We show the performance under unimodal settings by keeping only the image or text encoders and compare it with the multimodal model. }
\vspace{-2mm}
\centering
\resizebox{\linewidth}{!}{
    \begin{tabular}{c|cc|cc|cc}
    \hline
     \multicolumn{1}{c|}{\multirow{2}{*}{\textbf{Model}}} & \multicolumn{2}{c|}{\textbf{MVSA-Single}} & \multicolumn{2}{c|}{\textbf{MVSA-Multiple}} & \multicolumn{2}{c}{\textbf{TumEmo}} \\
          \multicolumn{1}{c|}{} & \textbf{Acc$\uparrow$} & \textbf{F1$\uparrow$} & \textbf{Acc$\uparrow$} & \textbf{F1$\uparrow$} & \textbf{Acc$\uparrow$} & \textbf{F1$\uparrow$} \\
    \hline 
    BERTweet & 72.73 & 70.87 & 69.84 & 64.18 & 67.90 & 67.58 \\
     + Pre-trained textual feature & 72.49 & 71.28 & 70.18 & 66.90 & 68.99 & 68.75 \\\hline
    Swin & 68.46 & 66.40 & 67.94 & 60.18 & 54.04 & 53.89 \\
     + Pre-trained visual features & 70.15 & 68.96 & 68.28 & 63.45 & 61.40 & 61.51 \\\hline
    BERTweet + Swin & 75.52 & 74.60 & 70.83 & 68.23 & 76.18 & 76.17 \\
    VSA-PF (ours) & 75.58 & 74.48 & 71.26 & 68.72 & 76.58 & 76.57 \\\hline
    \end{tabular}
    }
    \vspace{-6mm}
\label{table:ablation modality}
\end{table}

\nb{The Necessity of Multimodal Input.}
Table~\ref{table:ablation modality} shows that the pre-trained visual (CLIP image and visual expert features) and textual (CLIP text feature) elements enhance unimodal sentiment analysis over using Swin Transformer and BERTweet alone. This supports the effectiveness of pre-trained features.
When using unimodal data only, text-based analysis generally outperforms image-based analysis. This is \textit{in line with} previous findings~\cite{yang2021multimodal,li2022clmlf}, and could be due to a higher level of abstraction and subjectivity in the visual modality. 
Multimodal models show consistently superior performance than unimodal models, particularly on the challenging TumEmo dataset, which demonstrates the necessity of multimodal input for comprehensive sentiment analysis.

\nb{Fusion Module Architecture.}
We compare our BERT-based fusion module to a lightweight $3$-layer MLP with a $768$-size hidden layer, where MLP is used by~\cite{cheema2021fair} as the feature fusion module. In this setup, visual and textual features are concatenated and fed into the MLP with ReLU activation for sentiment prediction. Table~\ref{table:ablation branch} shows that MLP is comparable to BERT on the TumEmo but worse on two MVSA datasets, demonstrating that BERT is more effective in fusing multimodal multi-domain features than simple networks like MLP.

\section{Conclusion and Future Work}

In this paper, we present a holistic framework for visual-textual sentiment analysis, which consists of trainable visual-textual, visual-expert, and CLIP branches. The visual-textual branch exploits the power of state-of-the-art network architectures.
The visual expert branch allows our framework to benefit from rich pre-trained visual encoders. The CLIP branch equips the proposed method with the ability to model visual-textual correspondence explicitly. The experimental results on three public datasets demonstrate the effectiveness of our proposed framework across the board. The ablation study reveals the necessity of each designed branch. In the future, we plan to further investigate why some encoders are ineffective on certain datasets and address this limitation. Additionally, there is currently no systematic evaluation of foundational LLM/LVM models in multimodal sentiment analysis benchmarks. Our preliminary tests with GPT-4V on 30 samples from each dataset showed lower accuracies (76\%/66\%/43\%) compared to our framework, particularly on complex datasets. An interesting direction for future research is to integrate our method with LLMs and LVMs. For example, we could explore using embeddings extracted from these models or employ special adapters to integrate our pre-trained features into these models.

\clearpage

\bibliographystyle{IEEEbib}
\bibliography{report}

\clearpage

\end{document}